\documentclass[10pt, a4paper]{article}
\usepackage{lrec}
\usepackage[draft]{minted}
\usepackage{enumitem}
\setlist{noitemsep}
\usepackage{graphicx}
\usepackage{tabularx}
\usepackage{booktabs}
\usepackage{soul}
\usepackage{graphicx}
\usepackage{hyperref}
\usepackage{xstring}
\usepackage{tipa}
\usepackage{comment}
\usepackage{multirow}

\usepackage{tikz}
\usetikzlibrary{positioning,arrows}

\usepackage{moresize}

\makeatletter
\def\PYG@reset{\let\PYG@it=\relax \let\PYG@bf=\relax%
    \let\PYG@ul=\relax \let\PYG@tc=\relax%
    \let\PYG@bc=\relax \let\PYG@ff=\relax}
\def\PYG@tok#1{\csname PYG@tok@#1\endcsname}
\def\PYG@toks#1+{\ifx\relax#1\empty\else%
    \PYG@tok{#1}\expandafter\PYG@toks\fi}
\def\PYG@do#1{\PYG@bc{\PYG@tc{\PYG@ul{%
    \PYG@it{\PYG@bf{\PYG@ff{#1}}}}}}}
\def\PYG#1#2{\PYG@reset\PYG@toks#1+\relax+\PYG@do{#2}}

\expandafter\def\csname PYG@tok@gd\endcsname{\def\PYG@tc##1{\textcolor[rgb]{0.63,0.00,0.00}{##1}}}
\expandafter\def\csname PYG@tok@gu\endcsname{\let\PYG@bf=\textbf\def\PYG@tc##1{\textcolor[rgb]{0.50,0.00,0.50}{##1}}}
\expandafter\def\csname PYG@tok@gt\endcsname{\def\PYG@tc##1{\textcolor[rgb]{0.00,0.27,0.87}{##1}}}
\expandafter\def\csname PYG@tok@gs\endcsname{\let\PYG@bf=\textbf}
\expandafter\def\csname PYG@tok@gr\endcsname{\def\PYG@tc##1{\textcolor[rgb]{1.00,0.00,0.00}{##1}}}
\expandafter\def\csname PYG@tok@cm\endcsname{\let\PYG@it=\textit\def\PYG@tc##1{\textcolor[rgb]{0.25,0.50,0.50}{##1}}}
\expandafter\def\csname PYG@tok@vg\endcsname{\def\PYG@tc##1{\textcolor[rgb]{0.10,0.09,0.49}{##1}}}
\expandafter\def\csname PYG@tok@vi\endcsname{\def\PYG@tc##1{\textcolor[rgb]{0.10,0.09,0.49}{##1}}}
\expandafter\def\csname PYG@tok@vm\endcsname{\def\PYG@tc##1{\textcolor[rgb]{0.10,0.09,0.49}{##1}}}
\expandafter\def\csname PYG@tok@mh\endcsname{\def\PYG@tc##1{\textcolor[rgb]{0.40,0.40,0.40}{##1}}}
\expandafter\def\csname PYG@tok@cs\endcsname{\let\PYG@it=\textit\def\PYG@tc##1{\textcolor[rgb]{0.25,0.50,0.50}{##1}}}
\expandafter\def\csname PYG@tok@ge\endcsname{\let\PYG@it=\textit}
\expandafter\def\csname PYG@tok@vc\endcsname{\def\PYG@tc##1{\textcolor[rgb]{0.10,0.09,0.49}{##1}}}
\expandafter\def\csname PYG@tok@il\endcsname{\def\PYG@tc##1{\textcolor[rgb]{0.40,0.40,0.40}{##1}}}
\expandafter\def\csname PYG@tok@go\endcsname{\def\PYG@tc##1{\textcolor[rgb]{0.53,0.53,0.53}{##1}}}
\expandafter\def\csname PYG@tok@cp\endcsname{\def\PYG@tc##1{\textcolor[rgb]{0.74,0.48,0.00}{##1}}}
\expandafter\def\csname PYG@tok@gi\endcsname{\def\PYG@tc##1{\textcolor[rgb]{0.00,0.63,0.00}{##1}}}
\expandafter\def\csname PYG@tok@gh\endcsname{\let\PYG@bf=\textbf\def\PYG@tc##1{\textcolor[rgb]{0.00,0.00,0.50}{##1}}}
\expandafter\def\csname PYG@tok@ni\endcsname{\let\PYG@bf=\textbf\def\PYG@tc##1{\textcolor[rgb]{0.60,0.60,0.60}{##1}}}
\expandafter\def\csname PYG@tok@nl\endcsname{\def\PYG@tc##1{\textcolor[rgb]{0.63,0.63,0.00}{##1}}}
\expandafter\def\csname PYG@tok@nn\endcsname{\let\PYG@bf=\textbf\def\PYG@tc##1{\textcolor[rgb]{0.00,0.00,1.00}{##1}}}
\expandafter\def\csname PYG@tok@no\endcsname{\def\PYG@tc##1{\textcolor[rgb]{0.53,0.00,0.00}{##1}}}
\expandafter\def\csname PYG@tok@na\endcsname{\def\PYG@tc##1{\textcolor[rgb]{0.49,0.56,0.16}{##1}}}
\expandafter\def\csname PYG@tok@nb\endcsname{\def\PYG@tc##1{\textcolor[rgb]{0.00,0.50,0.00}{##1}}}
\expandafter\def\csname PYG@tok@nc\endcsname{\let\PYG@bf=\textbf\def\PYG@tc##1{\textcolor[rgb]{0.00,0.00,1.00}{##1}}}
\expandafter\def\csname PYG@tok@nd\endcsname{\def\PYG@tc##1{\textcolor[rgb]{0.67,0.13,1.00}{##1}}}
\expandafter\def\csname PYG@tok@ne\endcsname{\let\PYG@bf=\textbf\def\PYG@tc##1{\textcolor[rgb]{0.82,0.25,0.23}{##1}}}
\expandafter\def\csname PYG@tok@nf\endcsname{\def\PYG@tc##1{\textcolor[rgb]{0.00,0.00,1.00}{##1}}}
\expandafter\def\csname PYG@tok@si\endcsname{\let\PYG@bf=\textbf\def\PYG@tc##1{\textcolor[rgb]{0.73,0.40,0.53}{##1}}}
\expandafter\def\csname PYG@tok@s2\endcsname{\def\PYG@tc##1{\textcolor[rgb]{0.73,0.13,0.13}{##1}}}
\expandafter\def\csname PYG@tok@nt\endcsname{\let\PYG@bf=\textbf\def\PYG@tc##1{\textcolor[rgb]{0.00,0.50,0.00}{##1}}}
\expandafter\def\csname PYG@tok@nv\endcsname{\def\PYG@tc##1{\textcolor[rgb]{0.10,0.09,0.49}{##1}}}
\expandafter\def\csname PYG@tok@s1\endcsname{\def\PYG@tc##1{\textcolor[rgb]{0.73,0.13,0.13}{##1}}}
\expandafter\def\csname PYG@tok@dl\endcsname{\def\PYG@tc##1{\textcolor[rgb]{0.73,0.13,0.13}{##1}}}
\expandafter\def\csname PYG@tok@ch\endcsname{\let\PYG@it=\textit\def\PYG@tc##1{\textcolor[rgb]{0.25,0.50,0.50}{##1}}}
\expandafter\def\csname PYG@tok@m\endcsname{\def\PYG@tc##1{\textcolor[rgb]{0.40,0.40,0.40}{##1}}}
\expandafter\def\csname PYG@tok@gp\endcsname{\let\PYG@bf=\textbf\def\PYG@tc##1{\textcolor[rgb]{0.00,0.00,0.50}{##1}}}
\expandafter\def\csname PYG@tok@sh\endcsname{\def\PYG@tc##1{\textcolor[rgb]{0.73,0.13,0.13}{##1}}}
\expandafter\def\csname PYG@tok@ow\endcsname{\let\PYG@bf=\textbf\def\PYG@tc##1{\textcolor[rgb]{0.67,0.13,1.00}{##1}}}
\expandafter\def\csname PYG@tok@sx\endcsname{\def\PYG@tc##1{\textcolor[rgb]{0.00,0.50,0.00}{##1}}}
\expandafter\def\csname PYG@tok@bp\endcsname{\def\PYG@tc##1{\textcolor[rgb]{0.00,0.50,0.00}{##1}}}
\expandafter\def\csname PYG@tok@c1\endcsname{\let\PYG@it=\textit\def\PYG@tc##1{\textcolor[rgb]{0.25,0.50,0.50}{##1}}}
\expandafter\def\csname PYG@tok@fm\endcsname{\def\PYG@tc##1{\textcolor[rgb]{0.00,0.00,1.00}{##1}}}
\expandafter\def\csname PYG@tok@o\endcsname{\def\PYG@tc##1{\textcolor[rgb]{0.40,0.40,0.40}{##1}}}
\expandafter\def\csname PYG@tok@kc\endcsname{\let\PYG@bf=\textbf\def\PYG@tc##1{\textcolor[rgb]{0.00,0.50,0.00}{##1}}}
\expandafter\def\csname PYG@tok@c\endcsname{\let\PYG@it=\textit\def\PYG@tc##1{\textcolor[rgb]{0.25,0.50,0.50}{##1}}}
\expandafter\def\csname PYG@tok@mf\endcsname{\def\PYG@tc##1{\textcolor[rgb]{0.40,0.40,0.40}{##1}}}
\expandafter\def\csname PYG@tok@err\endcsname{\def\PYG@bc##1{\setlength{\fboxsep}{0pt}\fcolorbox[rgb]{1.00,0.00,0.00}{1,1,1}{\strut ##1}}}
\expandafter\def\csname PYG@tok@mb\endcsname{\def\PYG@tc##1{\textcolor[rgb]{0.40,0.40,0.40}{##1}}}
\expandafter\def\csname PYG@tok@ss\endcsname{\def\PYG@tc##1{\textcolor[rgb]{0.10,0.09,0.49}{##1}}}
\expandafter\def\csname PYG@tok@sr\endcsname{\def\PYG@tc##1{\textcolor[rgb]{0.73,0.40,0.53}{##1}}}
\expandafter\def\csname PYG@tok@mo\endcsname{\def\PYG@tc##1{\textcolor[rgb]{0.40,0.40,0.40}{##1}}}
\expandafter\def\csname PYG@tok@kd\endcsname{\let\PYG@bf=\textbf\def\PYG@tc##1{\textcolor[rgb]{0.00,0.50,0.00}{##1}}}
\expandafter\def\csname PYG@tok@mi\endcsname{\def\PYG@tc##1{\textcolor[rgb]{0.40,0.40,0.40}{##1}}}
\expandafter\def\csname PYG@tok@kn\endcsname{\let\PYG@bf=\textbf\def\PYG@tc##1{\textcolor[rgb]{0.00,0.50,0.00}{##1}}}
\expandafter\def\csname PYG@tok@cpf\endcsname{\let\PYG@it=\textit\def\PYG@tc##1{\textcolor[rgb]{0.25,0.50,0.50}{##1}}}
\expandafter\def\csname PYG@tok@kr\endcsname{\let\PYG@bf=\textbf\def\PYG@tc##1{\textcolor[rgb]{0.00,0.50,0.00}{##1}}}
\expandafter\def\csname PYG@tok@s\endcsname{\def\PYG@tc##1{\textcolor[rgb]{0.73,0.13,0.13}{##1}}}
\expandafter\def\csname PYG@tok@kp\endcsname{\def\PYG@tc##1{\textcolor[rgb]{0.00,0.50,0.00}{##1}}}
\expandafter\def\csname PYG@tok@w\endcsname{\def\PYG@tc##1{\textcolor[rgb]{0.73,0.73,0.73}{##1}}}
\expandafter\def\csname PYG@tok@kt\endcsname{\def\PYG@tc##1{\textcolor[rgb]{0.69,0.00,0.25}{##1}}}
\expandafter\def\csname PYG@tok@sc\endcsname{\def\PYG@tc##1{\textcolor[rgb]{0.73,0.13,0.13}{##1}}}
\expandafter\def\csname PYG@tok@sb\endcsname{\def\PYG@tc##1{\textcolor[rgb]{0.73,0.13,0.13}{##1}}}
\expandafter\def\csname PYG@tok@sa\endcsname{\def\PYG@tc##1{\textcolor[rgb]{0.73,0.13,0.13}{##1}}}
\expandafter\def\csname PYG@tok@k\endcsname{\let\PYG@bf=\textbf\def\PYG@tc##1{\textcolor[rgb]{0.00,0.50,0.00}{##1}}}
\expandafter\def\csname PYG@tok@se\endcsname{\let\PYG@bf=\textbf\def\PYG@tc##1{\textcolor[rgb]{0.73,0.40,0.13}{##1}}}
\expandafter\def\csname PYG@tok@sd\endcsname{\let\PYG@it=\textit\def\PYG@tc##1{\textcolor[rgb]{0.73,0.13,0.13}{##1}}}

\def\PYGZob{\char`\{}
\def\PYGZcb{\char`\}}

\def\PYGZhy{\char`\-}

\def\PYGZdq{\char`\"}

\makeatother

\makeatletter
\def\PYGdefault@reset{\let\PYGdefault@it=\relax \let\PYGdefault@bf=\relax%
    \let\PYGdefault@ul=\relax \let\PYGdefault@tc=\relax%
    \let\PYGdefault@bc=\relax \let\PYGdefault@ff=\relax}
\def\PYGdefault@tok#1{\csname PYGdefault@tok@#1\endcsname}
\def\PYGdefault@toks#1+{\ifx\relax#1\empty\else%
    \PYGdefault@tok{#1}\expandafter\PYGdefault@toks\fi}
\def\PYGdefault@do#1{\PYGdefault@bc{\PYGdefault@tc{\PYGdefault@ul{%
    \PYGdefault@it{\PYGdefault@bf{\PYGdefault@ff{#1}}}}}}}
\def\PYGdefault#1#2{\PYGdefault@reset\PYGdefault@toks#1+\relax+\PYGdefault@do{#2}}

\expandafter\def\csname PYGdefault@tok@gd\endcsname{\def\PYGdefault@tc##1{\textcolor[rgb]{0.63,0.00,0.00}{##1}}}
\expandafter\def\csname PYGdefault@tok@gu\endcsname{\let\PYGdefault@bf=\textbf\def\PYGdefault@tc##1{\textcolor[rgb]{0.50,0.00,0.50}{##1}}}
\expandafter\def\csname PYGdefault@tok@gt\endcsname{\def\PYGdefault@tc##1{\textcolor[rgb]{0.00,0.27,0.87}{##1}}}
\expandafter\def\csname PYGdefault@tok@gs\endcsname{\let\PYGdefault@bf=\textbf}
\expandafter\def\csname PYGdefault@tok@gr\endcsname{\def\PYGdefault@tc##1{\textcolor[rgb]{1.00,0.00,0.00}{##1}}}
\expandafter\def\csname PYGdefault@tok@cm\endcsname{\let\PYGdefault@it=\textit\def\PYGdefault@tc##1{\textcolor[rgb]{0.25,0.50,0.50}{##1}}}
\expandafter\def\csname PYGdefault@tok@vg\endcsname{\def\PYGdefault@tc##1{\textcolor[rgb]{0.10,0.09,0.49}{##1}}}
\expandafter\def\csname PYGdefault@tok@vi\endcsname{\def\PYGdefault@tc##1{\textcolor[rgb]{0.10,0.09,0.49}{##1}}}
\expandafter\def\csname PYGdefault@tok@vm\endcsname{\def\PYGdefault@tc##1{\textcolor[rgb]{0.10,0.09,0.49}{##1}}}
\expandafter\def\csname PYGdefault@tok@mh\endcsname{\def\PYGdefault@tc##1{\textcolor[rgb]{0.40,0.40,0.40}{##1}}}
\expandafter\def\csname PYGdefault@tok@cs\endcsname{\let\PYGdefault@it=\textit\def\PYGdefault@tc##1{\textcolor[rgb]{0.25,0.50,0.50}{##1}}}
\expandafter\def\csname PYGdefault@tok@ge\endcsname{\let\PYGdefault@it=\textit}
\expandafter\def\csname PYGdefault@tok@vc\endcsname{\def\PYGdefault@tc##1{\textcolor[rgb]{0.10,0.09,0.49}{##1}}}
\expandafter\def\csname PYGdefault@tok@il\endcsname{\def\PYGdefault@tc##1{\textcolor[rgb]{0.40,0.40,0.40}{##1}}}
\expandafter\def\csname PYGdefault@tok@go\endcsname{\def\PYGdefault@tc##1{\textcolor[rgb]{0.53,0.53,0.53}{##1}}}
\expandafter\def\csname PYGdefault@tok@cp\endcsname{\def\PYGdefault@tc##1{\textcolor[rgb]{0.74,0.48,0.00}{##1}}}
\expandafter\def\csname PYGdefault@tok@gi\endcsname{\def\PYGdefault@tc##1{\textcolor[rgb]{0.00,0.63,0.00}{##1}}}
\expandafter\def\csname PYGdefault@tok@gh\endcsname{\let\PYGdefault@bf=\textbf\def\PYGdefault@tc##1{\textcolor[rgb]{0.00,0.00,0.50}{##1}}}
\expandafter\def\csname PYGdefault@tok@ni\endcsname{\let\PYGdefault@bf=\textbf\def\PYGdefault@tc##1{\textcolor[rgb]{0.60,0.60,0.60}{##1}}}
\expandafter\def\csname PYGdefault@tok@nl\endcsname{\def\PYGdefault@tc##1{\textcolor[rgb]{0.63,0.63,0.00}{##1}}}
\expandafter\def\csname PYGdefault@tok@nn\endcsname{\let\PYGdefault@bf=\textbf\def\PYGdefault@tc##1{\textcolor[rgb]{0.00,0.00,1.00}{##1}}}
\expandafter\def\csname PYGdefault@tok@no\endcsname{\def\PYGdefault@tc##1{\textcolor[rgb]{0.53,0.00,0.00}{##1}}}
\expandafter\def\csname PYGdefault@tok@na\endcsname{\def\PYGdefault@tc##1{\textcolor[rgb]{0.49,0.56,0.16}{##1}}}
\expandafter\def\csname PYGdefault@tok@nb\endcsname{\def\PYGdefault@tc##1{\textcolor[rgb]{0.00,0.50,0.00}{##1}}}
\expandafter\def\csname PYGdefault@tok@nc\endcsname{\let\PYGdefault@bf=\textbf\def\PYGdefault@tc##1{\textcolor[rgb]{0.00,0.00,1.00}{##1}}}
\expandafter\def\csname PYGdefault@tok@nd\endcsname{\def\PYGdefault@tc##1{\textcolor[rgb]{0.67,0.13,1.00}{##1}}}
\expandafter\def\csname PYGdefault@tok@ne\endcsname{\let\PYGdefault@bf=\textbf\def\PYGdefault@tc##1{\textcolor[rgb]{0.82,0.25,0.23}{##1}}}
\expandafter\def\csname PYGdefault@tok@nf\endcsname{\def\PYGdefault@tc##1{\textcolor[rgb]{0.00,0.00,1.00}{##1}}}
\expandafter\def\csname PYGdefault@tok@si\endcsname{\let\PYGdefault@bf=\textbf\def\PYGdefault@tc##1{\textcolor[rgb]{0.73,0.40,0.53}{##1}}}
\expandafter\def\csname PYGdefault@tok@s2\endcsname{\def\PYGdefault@tc##1{\textcolor[rgb]{0.73,0.13,0.13}{##1}}}
\expandafter\def\csname PYGdefault@tok@nt\endcsname{\let\PYGdefault@bf=\textbf\def\PYGdefault@tc##1{\textcolor[rgb]{0.00,0.50,0.00}{##1}}}
\expandafter\def\csname PYGdefault@tok@nv\endcsname{\def\PYGdefault@tc##1{\textcolor[rgb]{0.10,0.09,0.49}{##1}}}
\expandafter\def\csname PYGdefault@tok@s1\endcsname{\def\PYGdefault@tc##1{\textcolor[rgb]{0.73,0.13,0.13}{##1}}}
\expandafter\def\csname PYGdefault@tok@dl\endcsname{\def\PYGdefault@tc##1{\textcolor[rgb]{0.73,0.13,0.13}{##1}}}
\expandafter\def\csname PYGdefault@tok@ch\endcsname{\let\PYGdefault@it=\textit\def\PYGdefault@tc##1{\textcolor[rgb]{0.25,0.50,0.50}{##1}}}
\expandafter\def\csname PYGdefault@tok@m\endcsname{\def\PYGdefault@tc##1{\textcolor[rgb]{0.40,0.40,0.40}{##1}}}
\expandafter\def\csname PYGdefault@tok@gp\endcsname{\let\PYGdefault@bf=\textbf\def\PYGdefault@tc##1{\textcolor[rgb]{0.00,0.00,0.50}{##1}}}
\expandafter\def\csname PYGdefault@tok@sh\endcsname{\def\PYGdefault@tc##1{\textcolor[rgb]{0.73,0.13,0.13}{##1}}}
\expandafter\def\csname PYGdefault@tok@ow\endcsname{\let\PYGdefault@bf=\textbf\def\PYGdefault@tc##1{\textcolor[rgb]{0.67,0.13,1.00}{##1}}}
\expandafter\def\csname PYGdefault@tok@sx\endcsname{\def\PYGdefault@tc##1{\textcolor[rgb]{0.00,0.50,0.00}{##1}}}
\expandafter\def\csname PYGdefault@tok@bp\endcsname{\def\PYGdefault@tc##1{\textcolor[rgb]{0.00,0.50,0.00}{##1}}}
\expandafter\def\csname PYGdefault@tok@c1\endcsname{\let\PYGdefault@it=\textit\def\PYGdefault@tc##1{\textcolor[rgb]{0.25,0.50,0.50}{##1}}}
\expandafter\def\csname PYGdefault@tok@fm\endcsname{\def\PYGdefault@tc##1{\textcolor[rgb]{0.00,0.00,1.00}{##1}}}
\expandafter\def\csname PYGdefault@tok@o\endcsname{\def\PYGdefault@tc##1{\textcolor[rgb]{0.40,0.40,0.40}{##1}}}
\expandafter\def\csname PYGdefault@tok@kc\endcsname{\let\PYGdefault@bf=\textbf\def\PYGdefault@tc##1{\textcolor[rgb]{0.00,0.50,0.00}{##1}}}
\expandafter\def\csname PYGdefault@tok@c\endcsname{\let\PYGdefault@it=\textit\def\PYGdefault@tc##1{\textcolor[rgb]{0.25,0.50,0.50}{##1}}}
\expandafter\def\csname PYGdefault@tok@mf\endcsname{\def\PYGdefault@tc##1{\textcolor[rgb]{0.40,0.40,0.40}{##1}}}
\expandafter\def\csname PYGdefault@tok@err\endcsname{\def\PYGdefault@bc##1{\setlength{\fboxsep}{0pt}\fcolorbox[rgb]{1.00,0.00,0.00}{1,1,1}{\strut ##1}}}
\expandafter\def\csname PYGdefault@tok@mb\endcsname{\def\PYGdefault@tc##1{\textcolor[rgb]{0.40,0.40,0.40}{##1}}}
\expandafter\def\csname PYGdefault@tok@ss\endcsname{\def\PYGdefault@tc##1{\textcolor[rgb]{0.10,0.09,0.49}{##1}}}
\expandafter\def\csname PYGdefault@tok@sr\endcsname{\def\PYGdefault@tc##1{\textcolor[rgb]{0.73,0.40,0.53}{##1}}}
\expandafter\def\csname PYGdefault@tok@mo\endcsname{\def\PYGdefault@tc##1{\textcolor[rgb]{0.40,0.40,0.40}{##1}}}
\expandafter\def\csname PYGdefault@tok@kd\endcsname{\let\PYGdefault@bf=\textbf\def\PYGdefault@tc##1{\textcolor[rgb]{0.00,0.50,0.00}{##1}}}
\expandafter\def\csname PYGdefault@tok@mi\endcsname{\def\PYGdefault@tc##1{\textcolor[rgb]{0.40,0.40,0.40}{##1}}}
\expandafter\def\csname PYGdefault@tok@kn\endcsname{\let\PYGdefault@bf=\textbf\def\PYGdefault@tc##1{\textcolor[rgb]{0.00,0.50,0.00}{##1}}}
\expandafter\def\csname PYGdefault@tok@cpf\endcsname{\let\PYGdefault@it=\textit\def\PYGdefault@tc##1{\textcolor[rgb]{0.25,0.50,0.50}{##1}}}
\expandafter\def\csname PYGdefault@tok@kr\endcsname{\let\PYGdefault@bf=\textbf\def\PYGdefault@tc##1{\textcolor[rgb]{0.00,0.50,0.00}{##1}}}
\expandafter\def\csname PYGdefault@tok@s\endcsname{\def\PYGdefault@tc##1{\textcolor[rgb]{0.73,0.13,0.13}{##1}}}
\expandafter\def\csname PYGdefault@tok@kp\endcsname{\def\PYGdefault@tc##1{\textcolor[rgb]{0.00,0.50,0.00}{##1}}}
\expandafter\def\csname PYGdefault@tok@w\endcsname{\def\PYGdefault@tc##1{\textcolor[rgb]{0.73,0.73,0.73}{##1}}}
\expandafter\def\csname PYGdefault@tok@kt\endcsname{\def\PYGdefault@tc##1{\textcolor[rgb]{0.69,0.00,0.25}{##1}}}
\expandafter\def\csname PYGdefault@tok@sc\endcsname{\def\PYGdefault@tc##1{\textcolor[rgb]{0.73,0.13,0.13}{##1}}}
\expandafter\def\csname PYGdefault@tok@sb\endcsname{\def\PYGdefault@tc##1{\textcolor[rgb]{0.73,0.13,0.13}{##1}}}
\expandafter\def\csname PYGdefault@tok@sa\endcsname{\def\PYGdefault@tc##1{\textcolor[rgb]{0.73,0.13,0.13}{##1}}}
\expandafter\def\csname PYGdefault@tok@k\endcsname{\let\PYGdefault@bf=\textbf\def\PYGdefault@tc##1{\textcolor[rgb]{0.00,0.50,0.00}{##1}}}
\expandafter\def\csname PYGdefault@tok@se\endcsname{\let\PYGdefault@bf=\textbf\def\PYGdefault@tc##1{\textcolor[rgb]{0.73,0.40,0.13}{##1}}}
\expandafter\def\csname PYGdefault@tok@sd\endcsname{\let\PYGdefault@it=\textit\def\PYGdefault@tc##1{\textcolor[rgb]{0.73,0.13,0.13}{##1}}}

\makeatother

\title{AlloVera: A Multilingual Allophone Database}

\name{David R. Mortensen$^*$, Xinjian Li$^*$, Patrick Littell$^\dagger$, Alexis Michaud$^\ddagger$, Shruti Rijhwani$^*$, \\\bf\large Antonios Anastasopoulos$^*$, Alan W. Black$^*$, Florian Metze$^*$, Graham Neubig$^*$}

\address{$^*$Carnegie Mellon University; $^\dagger$National Research Council of Canada; $^\ddagger$CNRS-LACITO \\
         $^*$5000 Forbes Ave, Pittsburgh PA 15213, USA;\\ $^\dagger$1200 Montreal Rd, Ottawa ON K1A0R6, Canada; $^\ddagger$7 rue Guy Môquet, 94800 Villejuif, France \\
         $^*$\{dmortens, xinjianl, srijhwan, aanastas, awb, fmetze, neubig\}@cs.cmu.edu;\\
         $^\dagger$patrick.littell@nrc-cnrc.gc.ca; $^\ddagger$alexis.michaud@cnrs.fr\\}

\abstract{We introduce a new resource, AlloVera, which provides mappings from 218 allophones to phonemes for 14 languages. Phonemes are contrastive phonological units, and allophones are their various concrete realizations, which are predictable from phonological context. While phonemic representations are language specific, phonetic representations (stated in terms of (allo)phones) are much closer to a universal (language-independent) transcription. AlloVera allows the training of speech recognition models that output phonetic transcriptions in the International Phonetic Alphabet (IPA), regardless of the input language. We show that a ``universal'' allophone model, Allosaurus, built with AlloVera, outperforms ``universal'' phonemic models and language-specific models on a speech-transcription task. We explore the implications of this technology (and related technologies) for the documentation of endangered and minority languages. We further explore other applications for which AlloVera will be suitable as it grows, including phonological typology.}

\begin{document}

\maketitleabstract

\section{Introduction}

Speech can be represented at various levels of abstraction \cite{Clark-Yallop-Fletcher:2007-introduction,Ladefoged-Johnson:2014-course}. It can be recorded as an acoustic signal or an articulatory score. It can be transcribed with a panoply of detail (a \textsc{narrow} transcription), or with less detail (\textsc{broad} transcription). In fact, it can be transcribed retaining only those features that are \textit{contrastive} within the language under description,
or with abstract symbols that stand for contrastive units. This latter mode of representation is what is called a \textsc{phonemic} representation while the finer-grained range of representations are \textsc{phonetic} representations. Most NLP technologies that represent speech through transcription do so at a phonemic level (that is, words are represented as strings of \textsc{phonemes}). For language-specific models and questions, such representations are often adequate and may even be preferable to the alternatives. However, in multilingual models, the language-specific nature of phonemic abstractions can be a liability. The added phonetic realism of even a broad phonetic representation moves transcriptions closer to a universal space where categories transcend the bounds of a particular language.

This paper describes AlloVera\footnote{\url{https://github.com/dmort27/allovera}}, a resource that maps between the phonemic representations produced by many NLP tools---including grapheme-to-phoneme (G2P) transducers like our own \cite{Mortensen-et-al:2018-epitran}---and broad phonetic representations. Specifically, it is a database of phoneme-allophone pairs (where an allophone is a phonetic realization of a phoneme---see \S~\ref{sec:phonemes-allophones} below) for 14 languages. It is designed for notational compatibility with existing G2P systems. The phonetic representations are relatively broad, a consequence of our sources, but they are phonetically realistic enough to improve performance on a speech-to-phone recognition task, as shown in \S~\ref{sec:experiments}

This resource has applications beyond universal speech-to-phone recognition, including approximate search and speech synthesis (in human language technologies) and phonetic/phonological typology (in linguistics). The usefulness of AlloVera for all purposes will increase as it grows to cover a broad range of the languages for which phonetic and phonological descriptions have been completed. However, to illustrate the usefulness of AlloVera, we will rely primarily on the zero-shot, universal ASR use-case in the evaluation in this paper.

\subsection{Phonemes and Allophones}
\label{sec:phonemes-allophones}

\begin{figure}[bh!]
  \centering
\begin{tikzpicture}[>=stealth, node distance=5mm and 1cm]
\node (peak_orth) {\itshape peak};
\node (speak_orth) [right = of peak_orth] {\itshape speak};
\node (ping_orth) [right = of speak_orth] {\itshape ping};
\node (bing_orth) [right = of ping_orth] {\itshape bing};
\node[xshift=1cm] (english) [above = 0.1\baselineskip of peak_orth] {\scshape English};
\node[xshift=1cm] (mandarin) [above = 0.1\baselineskip of ping_orth] {\scshape Mandarin Chinese};
\node (ping_gloss) [below = 0.1\baselineskip of ping_orth] {`level’};
\node (bing_gloss) [below = 0.1\baselineskip of bing_orth] {`ice’};
\node (peak_phoneme) [below = of peak_orth] {\textipa{/\textcolor{red}{p}ik/}};
\node (speak_phoneme) [below = of speak_orth] {\textipa{/s\textcolor{red}{p}ik/}};
\node (ping_phoneme) [below = of ping_orth] {\textipa{/\textcolor{red}{p\super h}iN/}};
\node (bing_phoneme) [below = of bing_orth] {\textipa{/\textcolor{red}{p}iŋ/}};
\node (peak_phone) [below = of peak_phoneme] {\textipa{[\textcolor{red}{p\super h}ik]}};
\node (speak_phone) [below = of speak_phoneme] {\textipa{[s\textcolor{red}{p}ik]}};
\node (ping_phone) [below = of ping_phoneme] {\textipa{[\textcolor{red}{p\super h}iN]}};
\node (bing_phone) [below = of bing_phoneme] {\textipa{[\textcolor{red}{p}iŋ]}};
\draw[<->] (peak_phoneme.south) to [bend right=10] (bing_phoneme.south);
\draw[<->] (speak_phoneme.south) to [bend right=10] (bing_phoneme.south);
\draw[<->] (peak_phone.south) to [bend right=10] (ping_phone.south);
\draw[<->] (speak_phone.south) to [bend right=10] (bing_phone.south);
\end{tikzpicture}
\caption{Words, phonemes (slashes), and phones (square brackets) in English and Mandarin Chinese}
\label{fig:phonemes-and-allophones}
\label{example}
\end{figure}
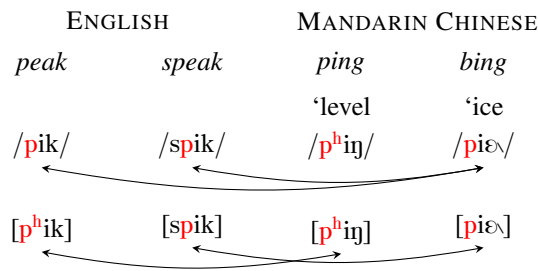

There have been various attempts at universal ASR: ``designing a universal phone recognizer which can decode a new target language with neither adaptation nor retraining'' \cite{siniscalchi2008toward}. This goal is up against major challenges. To begin with, defining the relevant units is no trivial task. Some research teams use grapheme-to-phoneme transducers to map orthography into a universal representational space. But in fact, as the name implies, these models typically yield \textit{phonemes} as their output and phonemes are, by their nature, language specific. Consider Figure~\ref{fig:phonemes-and-allophones}.

In English there is a single /\textipa{p}/ phoneme which is realized two different ways depending on the context in which it occurs \cite{Ladefoged:1999-american}. These contextual realizations are allophones. The aspirated allophone [\textipa{p\super h}] occurs word initially and at the beginning of stressed syllables. The unaspirated allophone [\textipa{p}] occurs in most other contexts. This is to be contrasted with Mandarin Chinese where there are distinct /\textipa{p\super h}/ and /\textipa{p}/ phonemes that are ``in contrast'' (that is, exchanging one for the other can result in a new morpheme) \cite{Norman:1988-chinese}. Mandarin /\textipa{p\super h}/ has one allophone, [\textipa{p\super h}], and Mandarin /\textipa{p}/ has one allophone, [\textipa{p}]. Thus, English and Mandarin have the same two ``p'' (allo)phones, but organize these into phonemes in different ways\footnote{In fact, the situation is even more complicated: [\textipa{b}], [\textipa{p}], and [\textipa{p\super h}] exist on a continuum called ``voice onset time'' or VOT. A sound transcribed as [\textipa{p}] in one language may have a voice onset time relatively close to [\textipa{p\super h}]. In another language, it may be similarly close to [\textipa{b}]. These categories, too, are to some degree language-specific \cite{abramson2017voice}. They are simply a step closer to phonetic reality than phonemic representations.}. 

\subsection{AlloVera and Multilingual Models}

For reasons stated above, multilingual training of speech models on English and Mandarin Chinese phonemes is problematic. A /\textipa{p\super h}/ phoneme in Chinese is always going to be roughly the same, phonetically, but a /\textipa{p}/ phoneme in English could be either [\textipa{p}] or [\textipa{p\super h}]. Once data from these two languages is combined, the contextual information separating the two sets of phones in English is erased and the result is a very noisy model. This problem is frequent when blending data from different languages. 

A thoroughly different way to go about doing multilingual training is defining a (universal) set of features to describe sounds in articulatory, acoustic or perceptual terms \cite{siniscalchi2008toward,johny2019cross}. But defining these features raises considerable epistemological difficulties.
There are cogent proposals from phoneticians and phonologists for transcribing sounds by means of their defining properties, rather than through International Phonetic Alphabet symbols \cite{vaissiere2011proposals}.
While these proposals appear promising in the mid/long run, they are not currently tractable to computational implementation in any straightforward way. Furthermore, the information that would be needed to implement such a system simply is not currently available to us in any form that we can consume.

The method explored here takes the middle ground: we create a database of allophones---that is to say, phonetic representations (referred to in phonetics/phonology as \textit{broad} phonetic transcriptions) rather than phonemic representations\footnote{An \textit{ideal} solution to such a problem might be to construct rule-based phoneme-to-allophone transducers (perhaps as FSTs) for each language in the training set. Then phonetic representations could be derived by first applying G2P to the orthographic text, then applying the appropriate transducer to the resulting phonemic representation. However, constructing such a resource is expensive, requires several specialized skills on the part of curators---who must encode the phonological environments in which allophones occur---and requires information that is often omitted from phonological descriptions of languages.}. This simplifies the annotation task: curators simply translate a set of relations among IPA symbols given in text to a simple allophone-phoneme mapping table. A curator can learn to do this with only a few hours of training. A multilingual model can then use these mappings in conjunction with speech data transcribed at the phonemic level to build a representation of each phone in the set.
This resource is described in the following sections.

\section{Description}

The AlloVera database is publicly available (via GitHub) at \url{https://github.com/dmort27/allovera} under an MIT license. In the following section, we explain the contents of the database, how they were curated, and give details about the data format and metadata provided for each language.

\subsection{Sources and Languages}

This resource consists of mappings between phonemic and broad phonetic representations for 14 languages with diverse phonologies. Languages were chosen based on three conditions:
\begin{itemize}
\item There is significant annotated speech data available for the language variety
\item There is an existing G2P system for the language variety or resources for adding support for that language variety to Epitran \cite{Mortensen-et-al:2018-epitran}.
\item There is a description of the phonology of the language variety including allophonic rules
\end{itemize}
The languages released in final form are listed in Table~\ref{tab:langs-in-allovera}. Additionally, there are currently several languages in alpha and beta states. Current alpha languages are Bengali (Indian; ben), Sundanese (sun), Swahili (swa), Portuguese (por), Cantonese (yue), Haitian (hat), and Zulu (zul). Current beta languages are Nepali (nep), Bengali (Bangladesh; ben), Korean (kor), Mongolian (mon), Greek (tsd), and Catalan (cat). We view AlloVera as an open-ended project which will continue expanding in the future.

\begin{table}[t]
  \footnotesize
  \centering
  \begin{tabularx}{1.0\linewidth}{p{1.5cm}ll>{\raggedright\arraybackslash}p{3.25cm}}
    \toprule
    \textbf{Language } & \textbf{Phonemes} & \textbf{Phones} & \textbf{Sources}\\
    \midrule
    Amharic & 49 & 57 & \newcite{Hayward-Hayward:1999-amharic}\\
    English (American) & 38 & 44 & \newcite{Ladefoged:1999-american}\\
    French & 36 & 38 & \newcite{Fougeron-Smith:1993-illustration}\\
    German & 40 & 42 & \newcite{Kunkel:2005-die-grammatik}\\
    Italian (Standard) & 41 & 45 & \newcite{Rogers-dArcangeli:2004-illustration}\\
    Japanese & 30 & 47 & \newcite{Wiki:2019-japanese-phonology}\\
    Javanese & 31 & 34 & \newcite[4--6]{Suharno:1982-descriptive}\\
    Kazakh & 41 & 45 & \newcite{McCollum-Chen:2018-illustration}\\
    Mandarin & 31 & 41 & \newcite{Norman:1988-chinese}\\
    Russian & 45 & 62 & \newcite{Yanushevskaya-Buncic:2015-illustration}\\
    Spanish & 30 & 39 & \newcite{Martinez-Celdran-et-al:2003-illustration,Wiki:2019-spanish-language}\\
    Tagalog & 29 & 42 & \newcite{Wiki:2019-tagalog-phonology}\\
    Turkish & 30 & 43 & \newcite{Zimmer-Orgun:1992-illustration}\\
    Vietnamese (Hanoi) & 34 & 42 & \newcite{Kirby:2011-illustration}\\
    \bottomrule
  \end{tabularx}
  \caption{Languages included in AlloVera}
  \label{tab:langs-in-allovera}
\end{table}

\subsection{Curation Practices}

Most mappings were initially encoded by non-experts with a few hours of training, but all were subsequently checked by the first author, a professional linguist with graduate training in phonetics and phonology.

Our policy, in creating the mappings, was to use---where available---the ``Illustrations of the IPA'' series published in the \textit{Handbook of the International Phonetic Association} \cite{IPA:1999-handbook} and the \textit{Journal of the International Phonetic Association} as our primary references. When that was not possible, we used other references, including Wikipedia summaries of research on the relevant languages. Each mapping was designed to be used with a particular G2P model. Curators mapped each phone in the description to the relevant phoneme using a spreadsheet. The phonemes from the standard (in IPA) were then mapped to the phonemes output by the G2P system (typically in X-SAMPA). When there was imperfect alignment between these sets, changes were typically made to the G2P model, expanding or restricting its range of outputs. However, in some cases, phonemes output by the G2P system could be shown to occur extra-systemically (for example, in loanwords) and the phoneme set was expanded to accommodate it. In these cases, we used equivalent IPA/X-SAMPA symbols for the phonemic and broad phonetic representations.

Languages show considerable internal variation. For example, the system of fricatives differs significantly between various varieties of Spanish. In some cases, our speech data is from a specific variety (e.g. Castilian Spanish). In other cases, it may be polydialectal. Where possible (as with Spanish), we have made the mappings general, so that they admit phoneme-allophone mappings from a variety of dialects. In other cases (as with German), however, our resources describe a single ``standard'' variety and the range of phonetic variation present in colloquial speech is not necessarily reflected in the mappings. Dialectal variation also posed a challenge when the datasets did not have sufficient information about the speakers and the dialect used in the recorded speech. In these cases, we resorted to asking native speakers to identify the appropriate variant and create the mappings based on their analysis. However, we observed that this task is sometimes difficult, even for life-long speakers of the language. For example, to differentiate between Indian and Bangladeshi variants of Bengali, multiple examples had to be presented to the native speakers in order to get a reasonably confident analysis of the dataset. In future work, we plan to increase the generality of as many of the mappings as possible by incorporating information from scholarly resources on phonetic variation in the relevant languages.

There were a few recurring challenges facing curators. These include descriptions that do not distinguish between allophonic and morphophonemic (=morphophonological) rules, or between allophonic rules and orthographic rules. In these cases, curators were told to ignore any abstraction above the level that would be produced by the G2P system. On a related front, some G2P systems---like the one we use for Japanese---generate archiphonemes, as with the Japanese moraic nasal \textit{N}. In these cases, we allowed an archiphonemic analysis even though it deviated from the phonemic ideal assumed by most of the mappings.

\subsection{Data format}

The data is distributed as a set of JSON files (one per language variety) and a BibTeX file containing source information. Each file contains the following metadata:
\begin{itemize}
\item The ISO 639-3 code for the language
\item The Glottocode(s) for the supported lect(s) \cite{glottolog}
\item The primary source for the mapping (as a BibTeX cite key)
\item The secondary sources for the mapping (as BibTeX cite keys)
\item If the mapping is constructed to be used with Epitran \cite{Mortensen-et-al:2018-epitran}, the associated Epitran language-script code.
\item If the mapping is made for use with another G2P engine, an identifier for this engine.
\end{itemize}
The data itself is represented as an array (rather than an object, in order to allow many-to-many mappings) Each element in this array has the following required fields:
\begin{itemize}
\item Phonemic representation (IPA)
\item Phonetic representation (IPA)
\end{itemize}
It may have the following optional fields:
\begin{itemize}
\item Environment (verbal description of the phonological environment in which an allophone occurs)
\item Source (when source for mapping differs from primary source)
\item Glottocodes, if the mapping only applies to a subset of the lects listed in the global metadata
\item Notes
\end{itemize}
An excerpt from one of these files is given in Figure~\ref{fig:json-spanish}.

\begin{figure}[tbh]
  \ssmall
  \centering
\begin{Verbatim}[commandchars=\\\{\}]
\PYG{p}{\PYGZob{}}
    \PYG{n+nt}{\PYGZdq{}iso\PYGZdq{}}\PYG{p}{:} \PYG{l+s+s2}{\PYGZdq{}spa\PYGZdq{}}\PYG{p}{,}
    \PYG{n+nt}{\PYGZdq{}glottocode\PYGZdq{}}\PYG{p}{:} \PYG{p}{[}
        \PYG{l+s+s2}{\PYGZdq{}amer1254\PYGZdq{}}\PYG{p}{,}
        \PYG{l+s+s2}{\PYGZdq{}cast1244\PYGZdq{}}
    \PYG{p}{],}
    \PYG{n+nt}{\PYGZdq{}primary src\PYGZdq{}}\PYG{p}{:} \PYG{l+s+s2}{\PYGZdq{}Martinez\PYGZhy{}Celdran\PYGZhy{}et\PYGZhy{}al:2003\PYGZhy{}illustration\PYGZdq{}}\PYG{p}{,}
    \PYG{n+nt}{\PYGZdq{}secondary srcs\PYGZdq{}}\PYG{p}{:} \PYG{p}{[}\PYG{l+s+s2}{\PYGZdq{}Wiki:2019\PYGZhy{}spanish\PYGZhy{}language\PYGZdq{}}\PYG{p}{],}
    \PYG{n+nt}{\PYGZdq{}epitran\PYGZdq{}}\PYG{p}{:} \PYG{l+s+s2}{\PYGZdq{}spa\PYGZhy{}Latn\PYGZdq{}}\PYG{p}{,}
    \PYG{n+nt}{\PYGZdq{}mappings\PYGZdq{}}\PYG{p}{:} \PYG{p}{[}
    \PYG{err}{...}
           \PYG{p}{\PYGZob{}}
            \PYG{n+nt}{\PYGZdq{}phone\PYGZdq{}}\PYG{p}{:} \PYG{l+s+s2}{\PYGZdq{}\textipa{X}\PYGZdq{}}\PYG{p}{,}
            \PYG{n+nt}{\PYGZdq{}phoneme\PYGZdq{}}\PYG{p}{:} \PYG{l+s+s2}{\PYGZdq{}x\PYGZdq{}}\PYG{p}{,}
            \PYG{n+nt}{\PYGZdq{}environment\PYGZdq{}}\PYG{p}{:} \PYG{l+s+s2}{\PYGZdq{}optionally, before a back vowel\PYGZdq{}}\PYG{p}{,}
            \PYG{n+nt}{\PYGZdq{}glottocodes\PYGZdq{}}\PYG{p}{:} \PYG{p}{[}
                \PYG{l+s+s2}{\PYGZdq{}cast1244\PYGZdq{}}
            \PYG{p}{]}
        \PYG{p}{\PYGZcb{},}
    \PYG{err}{...}
    \PYG{p}{]}
\PYG{p}{\PYGZcb{}}
\end{Verbatim}

\caption{Fragment of the JSON object for Spanish.}
\label{fig:json-spanish}
\end{figure}

\subsection{Summary of the Contents}

The database currently defines 218 phones which are associated with one of 148 phoneme symbols. This falls far short of the total number of phones in the languages of the world---the PHOIBLE database has 3,183 phones \cite{Phoible}---but AlloVera has good representation of the most common phones, as shown in Table~\ref{tab:top-phoible-phones}.

\begin{table}[htb]
  \centering
  \begin{tabularx}{1.0\linewidth}{ll}
    \toprule
    PHOIBLE Set & Intersection with AlloVera Phone Set\\
    \midrule
    Top 50 & 44\\
    Top 100 & 72\\
    Top 200 & 107\\
    \bottomrule
  \end{tabularx}
  \caption{Representation of top PHOIBLE phones (by number of languages) in AlloVera}
  \label{tab:top-phoible-phones}
\end{table}

\subsection{Limitations}

Currently, AlloVera does not support tone, stress or other suprasegmentals, despite including mappings for two tone languages (Mandarin Chinese and Vietnamese) and one language with a pitch accent or restricted tone system (Japanese), as well as several languages with contrastive stress (e.g. English). This is due, in large part, to the complexity of representing these phonemes---which are ``spread out'' over multiple segments---in terms of IPA strings. There are two separate standards for representing tone within IPA \cite{IPA:1999-handbook}, one of which is used primarily by linguists working on East and Southeast Asian languages (Chao tone letters written at the beginning or end of a syllable) and one of which is used by linguists working on languages elsewhere in the world (diacritics written over the nuclear vowel of a syllable). To achieve the multilingual aims of AlloVera, it would be necessary to have a single scheme for representing tone across languages. 

\section{Experiments}
\label{sec:experiments}

To highlight one application of AlloVera, we implement an allophone speech recognition system, Allosaurus. We compare its performance with standard universal phoneme model and language-specific model. The results suggest that Allovera helps to improve the phoneme error rate on both the training languages and two unseen languages.

\subsection{Multilingual Allophone Model}

\begin{figure}[t]
  \centering
  \includegraphics[width=\columnwidth]{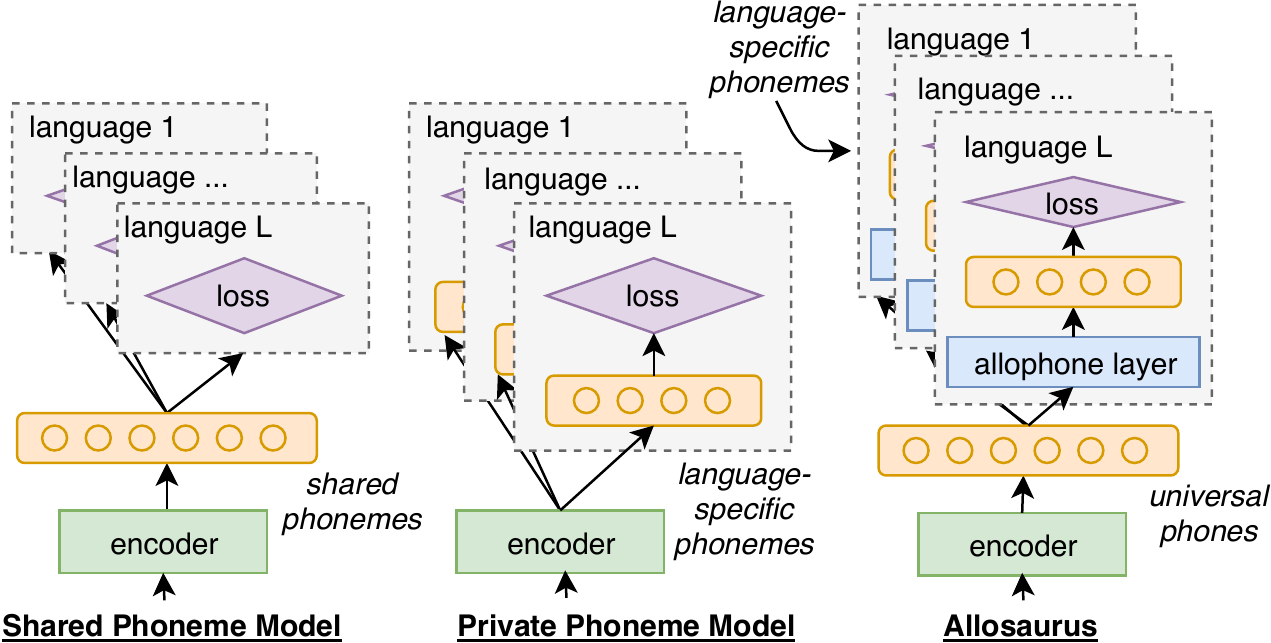}
  \caption{Traditional approaches predict phonemes directly, either for all languages (left) or separately for each language (middle). On the contrary, our approach (right) predicts over a shared phone inventory, then maps into language-specific phonemes with an allophone layer.}
\label{arch}
\end{figure}

The standard multilingual speech recognition models can be largely divided into two types as shown in Figure~\ref{arch}. The \textit{shared phoneme model} is a universal multilingual model which represents phonemes from all languages in a shared space. The underlying assumption of this architecture is that the same phoneme from different languages should be treated similarly in the acoustic model. This assumption is, however, not an very accurate approximation as the same phonemes from different languages are often realized by different allophones as described in \S~\ref{sec:phonemes-allophones}.

The second standard approach is the \textit{private phoneme model}, which is shown in the middle of the Figure~\ref{arch}. The model applies a language-specific classifier, which distinguishes the phonemes from different languages. This approach consists of a shared multilingual encoder and language-specific projection layer. This approach tends to perform better than the shared phoneme model, however, it fails to consider associations between phonemes across the languages. For example, /\textipa{p}/ in English and /\textipa{p}/ in Mandarin are treated as two completely distinct phonemes  despite the fact that their surface realizations overlap with each other. Additionally, it is difficult to derive language-independent phones or phonemes from this approach.

In contrast, the Allosaurus model described on the right side of Figure~\ref{arch} can overcome both issues of those standard models by taking advantage of AlloVera. Instead of constructing a shared phoneme set, the Allosaurus model constructs a shared phone set by taking the union of all 218 phones covered in the AlloVera dataset. The shared encoder first predicts the distribution over the phone set, then transforms the phone distribution into the phoneme distribution in each language using the allophone layer. The allophone layer is implemented by looking up the language-specific phone-phoneme correspondences as annotated in Allovera. By adopting this approach, the Allosaurus model overcomes the disadvantages of the two standard models: the phone representation is a much more appropriate choice for the language-independent representation than the shared phoneme representation, and this phone representation can be implemented without sacrificing language-specificity. For example, the language-independent phone [\textipa{p}] is first learned and then projected into English phoneme /\textipa{p}/ and Mandarin phoneme /\textipa{p}/.

\subsection{Results}
\begin{table*}[t!]
\begin{center}
    \begin{tabular}{l l | c c c c c c c c c c c | c} 
    \toprule
    & & Amh & Eng & Ger & Ita & Jap & Man & Rus & Spa & Tag & Tur & Vie & Average \\
    \midrule
    \multirow{3}{*}{\rotatebox{90}{\textbf{Full}}} & \textbf{Shared Phoneme PER}  & 78.4 & 71.7 & 71.6 & 62.9 & 65.9 & 76.5 & 76.9 & 62.6 & 74.1 & 76.6 & 82.7 & 73.8 \\
    & \textbf{Private Phoneme PER} & 37.1 & 22.4 & 17.6 & 26.2 & 17.6 & 17.9 & 21.3 & 18.5 & 47.6 & 35.8 & 56.5 & 25.6 \\
    & \textbf{Allosaurus PER} & 36.0 & 20.5 & 18.8 & 23.7 & 23.8 & 17.0 & 26.3 & 19.4 & 57.4 & 35.3 & 57.3 & 25.0 \\
    \midrule
    \multirow{3}{*}{\rotatebox{90}{\textbf{Low}}} & \textbf{Shared Phoneme PER} & 80.4 & 73.3 & 74.3 & 72.2  & 77.1 & 83.0 & 83.2 & 72.8 & 84.8 & 84.4 & 84.5 & 78.4 \\
    & \textbf{Private Phoneme PER} & 55.4 & 50.6 & 41.9 & 31.6 & 36.8 & 37.0 & 47.9 & 36.7 & 62.3 & 54.5 & 73.6 & 43.8 \\
    & \textbf{Allosaurus PER} & 54.8 & 47.0 & 41.5 & 37.4 & 40.5 & 33.4 & 45.0 & 35.9 & 70.1 & 53.6 & 72.5 & 41.8 \\
    \bottomrule
    \end{tabular}
    \caption{Three models' phoneme error rates on 11 languages. The top half shows the results when training with full datasets. The bottom half shows the low-resource results in which only 10k utterances are used for training from each dataset.} \label{tab:training_results} 
\end{center}
\end{table*}

To investigate how AlloVera improves multilingual speech recognition, we implemented three multilingual models mentioned above and compared their performance. In particular, we selected 11 languages from AlloVera taking into consideration the availability of those languages in our training speech corpus. For each language, we selected a training corpus from voxforge\footnote{\url{http://www.voxforge.org/}}, openSLR\footnote{\url{https://openslr.org/}} and other resources. The source and size of the data sets used in these experiments are given in Table~\ref{tab:corpus}.
\begin{table}[t!]
\begin{center}
    \resizebox{\columnwidth}{!}{
    \begin{tabular}{l p{2in} r} 
    \toprule
    {\bf Language} & {\bf Corpora} & {\bf Utt.} \\
    \midrule
    English & voxforge, Tedlium \cite{rousseau2012ted}, Switchboard \cite{godfrey1992switchboard} & 1148k \\
    Japanese & Japanese CSJ \cite{maekawa2003corpus} &  440k \\
    Mandarin & Hkust \cite{liu2006hkust}, openSLR \cite{aishell_2017,THCHS30_2015} & 377k \\
    Tagalog & IARPA-babel106b-v0.2g & 93k \\
    Turkish & IARPA-babel105b-v0.4 & 82k \\
    Vietnamese & IARPA-babel107b-v0.7 & 79k \\
    Kazakh & IARPA-babel302b-v1.0a & 48k \\
    German & voxforge & 40k \\
    Spanish & LDC2002S25 & 32k \\
    Amharic & openSLR25 \cite{Abate2005} & 10k \\
    Italian & voxforge & 10k \\
    Russian & voxforge & 8k \\
    \midrule
    Inukitut & private & 1k \\
    Tusom & private & 1k \\
    \bottomrule
    \end{tabular}
    }
\end{center}
    \caption{Training corpora and size in utterances for each language. Models are trained and tested with 12 rich resource languages (top) and 2 low resource unseen languages (bottom).} \label{tab:corpus}
\end{table}
To evaluate the model, we used 90\% of each corpus as the training set, and the remaining 10\% as the testing set. The evaluation metric is the phoneme error rate (PER) between the reference phonemes and hypothesis phonemes. For all three models, we applied the same bidirectional LSTM architecture as the encoder. The encoder has 6 layers and each layer has a hidden size of 1024. Additionally, the private phoneme model has a linear layer to map the hidden layer into language specific phoneme distributions and the Allosaurus model applies AlloVera to project the universal phone distribution into the language specific phoneme distributions. The loss function is CTC loss in all three models. The input features are 40-dimensional MFCCs.

Table~\ref{tab:training_results} show the performance of the three models under two different training conditions. The row tagged with Full means that the whole training set was used to train the multilingual model. In contrast, the row with tag Low is trained under a low resource condition in which we only select 10k utterances from each training corpus. This low resource condition is useful when building speech recognizers for new languages since training sets of most new languages are very limited. As Table~\ref{tab:training_results} suggests, the private phoneme model significantly outperforms the shared phoneme on all languages---the average PER of the shared phoneme model is 73.8\% and the private phoneme model has 25.6\% PER in the full training condition. During the evaluation process, we find that the performance of the shared phoneme model decreases significantly when increasing the number of training languages. This can be explained by the fact that phoneme assignment schemes are different across languages. Therefore, adding more languages can confuse the model, leading it to assign incorrect phonemes. In contrast, AlloVera provides a consistent assignment across languages by using allophone inventories. Comparing Allosaurus and the private phoneme model, we find that Allosaurus further improves from the private phoneme model by 0.6\% under the full condition and 2.0\% under the low resource condition. While the improvement is relatively limited in the full training case, it suggests AlloVera would be valuable for creating speech recognition models for low resource languages.

AlloVera gives Allosaurus another important capability---the ability to generate phones from the universal phone inventory. As Figure~\ref{arch} shows, the layer before the allophone layer represents the distribution over universal phone inventory. The universal phone inventory consists of all allophones in AlloVera. In contrast, the shared phoneme model could only generate inconsistent universal phonemes and the private phoneme model could only generate language-specific phonemes. Table~\ref{tab:unseen_results} highlights the generalization ability of Allosaurus and AlloVera over two unseen languages: Inuktitut and Tusom. The table suggests that Allosaurus and AlloVera improve the performance over both the shared phoneme model and the private phoneme model substantially.

\begin{table}[t]
\begin{center}
    \begin{tabular}{l r r} 
    \toprule
    & Inuktitut & Tusom \\
    \midrule
    \textbf{Shared Phoneme PER} & 94.1 & 93.5 \\
    \textbf{Private Phoneme PER} & 86.2 & 85.8 \\
    \textbf{Allosaurus PER} & 73.1 & 64.2 \\
    \bottomrule
    \end{tabular}
\end{center}
\caption{Comparisons of phone error rates in two unseen languages}
\label{tab:unseen_results}
\end{table}

\section{Applications}

Currently, we intend to integrate AlloVera and Allosaurus (or other future systems trained using AlloVera) into three practical downstream systems for very-low-resource languages, addressing tasks identified as development priorities in recent surveys of indigenous and other low-resource language technology \cite{thieberger2016doc,levow2017streamlined,littell:2018-indigenous}. 

In our experience, the most requested speech technology for very-low-resource languages is \textbf{transcription acceleration}, an application of speech recognition for decreasing the workload of transcribers.  Many low-resource and endangered languages do already have extensive \emph{untranscribed} speech collections, in the form of recorded radio broadcasts, linguists' field recordings, or other personal recordings.  Transcribing these collections is a high priority for many speech communities, as an untranscribed corpus is difficult to use in either research or education \cite{adams2018evaluating,foley2019elpis}.  AlloVera and Allosaurus were originally and primarily intended for use in transcription acceleration, although we will also be exploring other practical applications.

Another priority technology is \textbf{approximate search} of speech databases. While the aforementioned untranscribed speech collections can straightforwardly be made \emph{available} online, they are not especially \emph{accessible} as such. A researcher, teacher, or student cannot in practice listen to years' worth of radio recordings in search of a particular word or topic. AlloVera and Allosaurus, by making an approximate text representation of the corpus, open up the possibility for efficient approximate phonetic search through otherwise-untranscribed speech databases. Previous work has demonstrated the feasibility of such an approach~\cite{anastasopoulos-EtAl:2017:Speech-Centric,boito2017unwritten}, but the quality of the search results can be significantly boosted by improvements in a first-pass phonetic transcription~\cite{ondel2018bayesian}.

We are also planning on integrating AlloVera and Allosaurus into a language-neutral \textbf{forced-alignment} pipeline.  While forced-alignment is a task that is already commonly done in a zero-shot scenario (by manually mapping target-language phones to the vocabulary of a pretrained acoustic model, often an English one), the extensive phonetic vocabulary of AlloVera means that many phones 
are already covered.  This greatly expands the number of languages that can be aligned without the need for an extensive transcribed corpus or manual system configuration.

\section{Related Work}

AlloVera builds on work in three major areas: phonetics and theoretical phonology, phonological ontologies, and human language technologies.

The term \textsc{allophone} was coined by Benjamin Lee Whorf in the 1920s and was popularized by \newcite{Trager-Bloch:1941-syllabic}. However, the idea goes back much further, to \newcite{Baudoin-deCourtenay:1894}. The idea of allophony most relevant to our work here comes from American Structuralist linguists like \newcite{Harris:1951-structural}, but we also invoke the concept of the archiphoneme, associated with the Prague Circle \cite{Trubetskoy:1939-grundzuge}. In the 1950s and 1960s, the structuralist notions of the ``taxonomic'' phoneme and of allophones came under attack by generative linguists \cite{Chomsky-Halle:1968-sound,Halle:1962-phonology,Halle:1959-sound}, but they have retained their importance both in linguistic practice and linguistic theory.

Various resources containing phonological information, especially phonological inventories, have been compiled. An early resource was UCLA's UPSID. A more recent resource that combines UPSID and other segment inventories in a unified ontology is PHOIBLE \cite{Phoible}. However, due to the nature of PHOIBLE's sources, it is not always clear what level of representation is intended within a segment inventory and PHOIBLE does not consistently establish relationships between abstract segments---phonemes---and concrete segments---(allo)phones. In these respects, it is complementary to AlloVera.

\section{Conclusion}

AlloVera embraces the fact that it is useful to analyze the sounds of language at different levels. It allows scientists and engineers to build models that are based on phones using tools that generate phonemic representations. It also captures allophonic relations in a way that is more generally useful but which does not require a highly specialized notation (for example, for stating phonological environments). We have demonstrated its usefulness, in its current form, for a valuable task (zero-shot speech-to-phone recognition).

The resource will become even more useful as more languages are added. What we have produced so far should be seen as a proof of concept. As we develop the resource further, the accuracy of our recognizers will go up, our approximate search and forced alignment models will improve,  and new avenues of research will be opened. 

Achieving this goal will require participation from more than just our research team: we invite linguists and language scientists who have special knowledge or a particular interest in a language to contribute their knowledge to AlloVera in the form of a simple allophone-to-phoneme mapping (preferably with natural language descriptions of the environments in which each allophone occurs). With international participation, AlloVera can go from a database that is merely useful to a resource that is indispensable for speech and language research.

\section{Acknowledgements}

This   project   was   sponsored   in part by   the   Defense   Advanced Research  Projects  Agency  (DARPA)  Information  Innovation Office (I2O), program: Low Resource Languages for Emergent Incidents (LORELEI), issued by DARPA/I2O under Contract No. HR0011-15-C-0114. This material is also based in part upon work supported by the National Science Foundation under grant 1761548. Shruti Rijhwani is supported by a Bloomberg Data Science Ph.D. Fellowship. Alexis Michaud gratefully acknowledges grants ANR-10-LABX-0083 and ANR-19-CE38-0015.

\section{Bibliographical References}
\label{main:ref}

\bibliographystyle{lrec}
\bibliography{allophones}

\end{document}